\documentclass[10pt,twocolumn,letterpaper]{article}

\usepackage{cvpr}              

\usepackage{multirow}
\usepackage{booktabs}
\usepackage{graphicx}
\usepackage{makecell}

\usepackage{mathrsfs}
\usepackage{amssymb}

\usepackage{algorithm}
\usepackage{algorithmic}

\usepackage[most]{tcolorbox}
\usepackage{listings}
\usepackage{float}

\tcbset{
  aibox/.style={
    width=390pt,
    top=10pt,
    colback=white,
    colframe=black,
    colbacktitle=black,
    enhanced,
    center,
    attach boxed title to top left={yshift=-0.1in,xshift=0.15in},
    boxed title style={boxrule=0pt,colframe=white,},
  }
}
\newtcolorbox{AIbox}[2][]{aibox,title=#2,#1}

\definecolor{cvprblue}{rgb}{0.21,0.49,0.74}
\usepackage[pagebackref,breaklinks,colorlinks,allcolors=cvprblue]{hyperref}

\def\paperID{} 
\def\confName{CVPR}
\def\confYear{2026}

\title{Retrieving Counterfactuals Improves Visual In-Context Learning}

\author{Guangzhi Xiong \quad Sanchit Sinha \quad Zhenghao He \quad Aidong Zhang\\
University of Virginia
}

\begin{document}
\maketitle
\begin{abstract}
Vision-language models (VLMs) have achieved impressive performance across a wide range of multimodal reasoning tasks, but they often struggle to disentangle fine-grained visual attributes and reason about underlying causal relationships. In-context learning (ICL) offers a promising avenue for VLMs to adapt to new tasks, but its effectiveness critically depends on the selection of demonstration examples. Existing retrieval-augmented approaches typically rely on passive similarity-based retrieval, which tends to select correlated but non-causal examples, amplifying spurious associations and limiting model robustness. We introduce CIRCLES (Composed Image Retrieval for Causal Learning Example Selection), a novel framework that actively constructs demonstration sets by retrieving counterfactual-style examples through targeted, attribute-guided composed image retrieval. By incorporating counterfactual-style examples, CIRCLES enables VLMs to implicitly reason about the causal relations between attributes and outcomes, moving beyond superficial correlations and fostering more robust and grounded reasoning. Comprehensive experiments on four diverse datasets demonstrate that CIRCLES consistently outperforms existing methods across multiple architectures, especially on small-scale models, with pronounced gains under information scarcity. Furthermore, CIRCLES retrieves more diverse and causally informative examples, providing qualitative insights into how models leverage in-context demonstrations for improved reasoning. Our code is available at \url{https://github.com/gzxiong/CIRCLES}.
\end{abstract}    
\section{Introduction}
\label{sec:intro}

Vision-language models (VLMs) have recently demonstrated remarkable generalization across a wide spectrum of visual reasoning tasks, achieving impressive results on tasks such as visual question answering \cite{antol2015vqa,wu2017visual,du2023zero}, image captioning \cite{zhou2020unified,bucciarelli2024personalizing}, and multimodal classification \cite{saha2024improved,cooper2025rethinking,abdelhamed2024you}. 
This progress is largely attributed to their rich multimodal representations learned through large-scale pre-training.
However, in practice, VLMs often struggle with fine-grained and attribute-sensitive visual reasoning, settings prevalent in real-world vision applications \cite{kim2024finer,chen2024cello,hsieh2023sugarcrepe}.

To address these challenges, in-context learning (ICL) has emerged as a powerful paradigm, enabling VLMs to rapidly adapt to new tasks by conditioning on a small set of demonstration examples provided at test time \cite{alayrac2022flamingo,tsimpoukelli2021multimodal,zhou2024visual}. Recent work in visual ICL has highlighted the crucial role of retrieval-based selection strategies for constructing effective demonstration sets \cite{liu2022makes,chen2024understanding,chen2025provoking}. Standard approaches, such as RICES (Retrieval-based In-Context Example Selection) \cite{yang2022empirical,alayrac2022flamingo}, assemble in-context examples by identifying the nearest neighbors to the query image within the embedding space. Follow-up works such as MUIER \cite{luo2024does} and MMICES \cite{chen2025can} augment this process by leveraging multimodal similarity metrics that jointly consider both visual and textual features, enhancing the relevance of retrieved examples to the query.

Despite these advances, such similarity-oriented selection can lead to systematic errors when critical attributes are entangled with irrelevant co-occurrences. Because these retrieval baselines do not explicitly control or intervene on causal factors, models prompted with their selected demonstrations often learn to mimic surface correlations rather than identify the attributes that truly determine the answer. This limitation becomes more pronounced under information scarcity or distribution shifts, where relying solely on correlated neighbors offers little guidance on how changing specific attributes affects the outcome. These observations motivate approaches that go beyond similarity and instead curate demonstration sets that expose disentangled, attribute-level variations.

To address the limitations above, we introduce \textbf{CIRCLES} (Composed Image Retrieval for Causal Learning Example Selection), a new framework for visual in-context learning that explicitly incorporates counterfactual reasoning signals into the example selection process. Unlike prior work that selects examples only by image similarity, CIRCLES constructs demonstration sets by explicitly collecting attribute-level counterfactual-style examples via composed image retrieval (CIR), and combining them with standard similarity-based examples to create an informative context.

\begin{figure}[h!]
    \centering
    \includegraphics[width=1.0\linewidth]{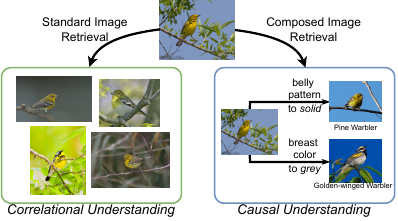}
    \caption{Illustration of how composed image retrieval provides additional causal understanding.}
    \label{fig:illustration}
\end{figure}

As illustrated in Figure \ref{fig:illustration}, standard similarity-based retrieval provides examples that are visually close to the query but may share irrelevant or confounding attributes. In contrast, by retrieving and composing images that reflect controlled interventions on key attributes (e.g., changing the belly pattern to solid while keeping other attributes as similar as possible), CIR identifies examples that isolate the effect of each attribute on the answer (e.g., similar images with a solid belly pattern are labeled as Pine Warbler), providing more causally informative demonstrations for ICL.

By composing causally informative examples with similarity-based demonstrations, CIRCLES constructs demonstration sets that better reveal the underlying factors driving the correct answer, enabling VLMs to learn disentangled, robust, and interpretable reasoning strategies. Our experiments on four benchmark datasets demonstrate that CIRCLES consistently outperforms existing ICL methods across a range of VLM architectures, with pronounced improvements on small-scale VLMs where models' internal knowledge is limited. CIRCLES is especially effective under challenging conditions such as information scarcity, where the performance gap widens as relevant data becomes limited. Qualitative analysis further shows that CIRCLES retrieves more diverse and informative examples that clarify critical attributes, providing insights into the ICL process.
Our main contributions are as follows:
\begin{itemize}
    \item We propose CIRCLES, a novel ICL framework that enriches demonstration sets by retrieving counterfactual examples, moving beyond standard similarity-based retrieval.
    \item Empirical evaluations on multiple image classification and visual question answering datasets demonstrate consistent improvements over existing ICL methods.
    \item Additional analyses highlight the strength of CIRCLES in low-data regimes and explore optimal practices for constructing informative demonstration sets.
\end{itemize}

\section{Related Work}

\subsection{Multimodal Reasoning}

Vision-language models (VLMs) have rapidly advanced visual tasks such as visual question answering (VQA), image captioning, and multimodal classification by integrating powerful visual encoders with large language models \cite{alayrac2022flamingo,li2023blip,liu2023visual}. Despite these gains, several studies have highlighted the limitations of VLMs in compositional and causal reasoning, revealing that state-of-the-art systems often rely on dataset priors or spurious correlations, rather than truly understanding fine-grained attributes or relational structure \cite{johnson2017clevr,agrawal2018don,sinha2025coco}. For instance, models can achieve high average scores but struggle when asked to reason about the effect of specific attribute changes or to generalize to out-of-distribution queries \cite{hudson2019gqa}. These observations motivate methods that structure inference-time evidence around attributes and relations rather than surface similarity.

\subsection{Visual In-Context Learning}

In-context learning (ICL) enables VLMs to adapt at inference time via conditioning on a small set of demonstration examples \cite{alayrac2022flamingo,tsimpoukelli2021multimodal,chen2025enhancing}, but the effectiveness of this process is highly dependent on the structure of the demonstration set \cite{liu2022makes,lu2022fantastically,rubin2022learning}. Recent research has shown the sensitivity of VLMs to the order, diversity, and retrieval strategies used to assemble demonstrations \cite{li2024configure,chen2024understanding,huang2024multimodal,yi2025drum}, demonstrating that naive similarity-based selection can lead to inconsistent performance, sensitivity to confounders, and susceptibility to spurious correlations \cite{balazevic2023towards,zhou2024explore,harutyunyan2024incontext}. 
These findings motivate moving beyond nearest-neighbor similarity toward example sets that are deliberately diverse and task-aligned.
Our approach complements this literature by explicitly introducing counterfactual examples into the demonstration pool, thus enriching the reasoning signal available to the model.

\subsection{Composed Image Retrieval}

Composed image retrieval (CIR) is a closely related area that focuses on retrieving images matching complex, compositionally specified queries, typically combining a reference image and a manipulation text \cite{song2025comprehensive,du2025survey}. Existing CIR techniques are optimized for retrieval accuracy rather than downstream reasoning or ICL composition \cite{vo2019composing,anwaar2021compositional}. Datasets such as FashionIQ \cite{wu2021fashion} and CIRR \cite{liu2021image} have advanced research on controlled attribute modifications, but these works treat CIR as an end task. Our use of CIR diverges from this paradigm: instead of retrieving a single modified image, we leverage CIR to construct a set of counterfactually informative demonstrations for in-context learning. By integrating attribute-guided text modifications with VLM-driven captioning, our method operationalizes CIR as a tool for counterfactual intervention, revealing which attributes meaningfully influence downstream predictions.

\begin{figure*}
    \centering
    \includegraphics[width=1.0\linewidth]{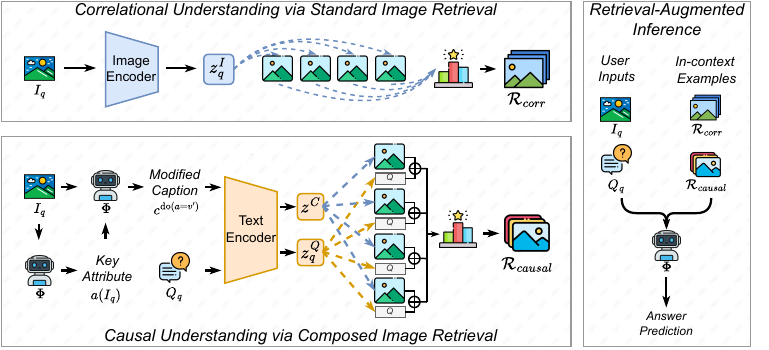}
    \caption{Overview of the CIRCLES framework. Given a query image \(I_q\) and question \(Q_q\), the top branch illustrates correlational understanding via standard image retrieval, while the bottom branch depicts causal understanding using attribute-guided composed image retrieval with counterfactual captions from the VLM \(\Phi\). Blue/orange rounded rectangles represent image/text embeddings. \(\mathcal{R}_{\text{corr}}\) and \(\mathcal{R}_{\text{causal}}\) denote the retrieved in-context examples for answer prediction.}

    \label{fig:architecture}
\end{figure*}

\section{Methodology} \label{sec:method}

In this section, we introduce the CIRCLES (Composed Image Retrieval for Causal Learning Example Selection) framework, designed to enhance visual in-context learning for vision-language models (VLMs). CIRCLES does not aim to perform formal causal identification. Instead, it leverages composed image retrieval (CIR) to approximate interventions on key attributes and to facilitate implicit in-context reasoning through contrastive retrieved examples. We begin by formalizing the problem and establishing notation. We then detail the CIRCLES pipeline, which consists of three main components: (1) causal understanding through attribute-guided CIR, (2) correlational understanding via standard image retrieval, and (3) retrieval-augmented inference.

\subsection{Task Formulation and Notation}

Let $\mathcal{K}=\{(I_j,Q_j,A_j)\}_{j=1}^N$ denote a visual question answering corpus, where $I_j$ is an image, $Q_j$ a natural language question, and $A_j$ the answer. Given a query $(I_q, Q_q)$, a VLM $\Phi$ produces an answer $A_q$ conditioned on the query and an optional retrieved context $\mathcal{R}\subseteq\mathcal{K}$.

We use a frozen CLIP model $E=(f_I,f_T)$ with image encoder $f_I$ and text encoder $f_T$ to obtain L2-normalized embeddings. For each candidate in the training corpus, we precompute embeddings $\mathbf{z}^I_j=f_I(I_j)$ and $\mathbf{z}^Q_j=f_T(Q_j)$. At test time, we compute $\mathbf{z}^I_q=f_I(I_q)$ and $\mathbf{z}^Q_q=f_T(Q_q)$ for the query. An overview of our CIRCLES framework is illustrated in Figure \ref{fig:architecture}.

\subsection{Causal Understanding via Attribute-Guided Composed Image Retrieval}

To collect demonstrations for causal understanding, CIRCLES identifies semantically meaningful attributes and retrieves counterfactual examples through a two-stage process.

\paragraph{Key Attribute Identification.}

Given $(I_q,Q_q)$, we prompt the VLM $\Phi$ to extract decisive attribute-value pairs for answering $Q_q$ (e.g., the attribute ``breast color'' has the value ``grey'' in Figure \ref{fig:illustration}). Let the set of attributes be 
\begin{equation}
{\mathcal{A}}=\{a_1,\dots,a_m\},
\end{equation}
with corresponding values on $I_q$ as
\begin{equation}
    \mathbf{v} = (v_1,\dots,v_m), \quad v_i = a_i(I_q).
\end{equation}
Each attribute $a_i$ has a finite value set $\mathcal{V}_i$ (e.g., possible colors for ``breast color''). These attributes form the basis for counterfactual intervention.

\paragraph{Counterfactual Example Retrieval.}

To isolate the influence of each attribute on the answer, we consider counterfactual variants of the query image by changing one attribute $a_i$ from its original value $v_i$ to an alternative $v_i' \in \mathcal{V}_i \setminus \{v_i\}$ while keeping all other attributes fixed. Direct identification of such counterfactuals in real datasets is generally infeasible. Therefore, for each attribute $a_i$ and candidate counterfactual value $v_i'$, we prompt the VLM to generate a counterfactual caption $c^{\text{do}(a_i = v_i')}$ which describes $I_q$ under the atomic intervention $do(a_i = v_i')$ \cite{pearl2009causality} with only attribute $a_i$ set to $v_i'$ and all $a_{k \neq i}$ unchanged.

For each candidate example $(I_j, Q_j, A_j)$ in $\mathcal{K}$, we compute the image-caption similarity score
\begin{equation}
    s^{\text{img}}_j = \mathbf{z}^{I\top}_j f_T(c^{\text{do}(a_i = v_i')}),
\end{equation}
which measures the visual faithfulness of $I_j$ to the counterfactual scenario described by $c^{\text{do}(a_i = v_i')}$.

However, relying solely on image-caption similarity can result in retrieving examples that visually align with the counterfactual but are not semantically relevant to the original query, particularly when attribute changes alter the context in unintended ways. To address this, we introduce an additional question-question similarity score
\begin{equation}\label{eq:task_similarity}
    s^{\text{txt}}_j = \mathbf{z}^{Q\top}_q \mathbf{z}^Q_j,
\end{equation}
which serves as a semantic constraint to ensure that retrieved examples not only match the counterfactual attributes but also remain closely related to the original question context.

The score for each candidate is computed as
\begin{equation}
    S_j = s^{\text{img}}_j + s^{\text{txt}}_j.
\end{equation}
By combining both visual faithfulness and semantic relevance, this scoring function identifies training examples that best approximate the intended counterfactual scenario for the given query.

All candidates are ranked according to $S_j$, and the top-$k_{\text{causal}}$ examples are selected to form the set
\begin{equation}
    \mathcal{R}^{\text{comp}}_i(v_i\!\to\! v_i') = \operatorname{TopK}_{j}(S_j),
\end{equation}
where $\operatorname{TopK}_{j}$ returns the top examples by the scores.
Aggregating across attributes and sampled counterfactuals yields the causal retrieval pool:
\begin{equation}
    \mathcal{R}_{\text{causal}} = \bigcup_{i=1}^m \mathcal{R}^{\text{comp}}_i(v_i\!\to\! \tilde v_i'),
\end{equation}
where $\tilde v_i'$ is one alternative value sampled from $\mathcal{V}_i \setminus \{v_i\}$ for efficiency, especially when $\mathcal{V}_i$ is large. 
In practice, we implement this sampling by prompting the VLM to suggest a plausible alternative value for each attribute.

\subsection{Correlational Understanding via Standard Image Retrieval}

To complement causal understanding given by the counterfactual examples, CIRCLES incorporates a correlation-oriented retrieval stage that provides broader contextual information. Here, we retrieve images that are most similar to the query image in the embedding space, without enforcing any attribute-based intervention or counterfactual modification. This approach helps the model leverage common visual or semantic patterns present in the dataset, supporting recognition and grounding even when explicit causal factors are absent. 

For query image $I_q$, we compute the image-image similarity score for all candidates in $\mathcal{K}$ as
\begin{equation}
    s^{\text{corr}}_j = \mathbf{z}^{I\top}_q \mathbf{z}^I_j,
\end{equation}
and select the top-$k_{\text{corr}}$ most similar examples to form the correlation retrieval set
\begin{equation}
    \mathcal{R}_{\text{corr}} = \operatorname{TopK}_{j}(s^{\text{corr}}_j).
\end{equation}
While advanced retrieval methods exist \cite{luo2024does,chen2025can}, we adopt this standard image-only retrieval for its efficiency and generality, which matches the standard RICES-style baseline \cite{yang2022empirical,alayrac2022flamingo} and cleanly isolates the added value of the proposed causal understanding module.
Our experiments in Appendix \ref{app:ablation_design} further illustrate that image-only retrieval is sufficient when the task is dominated by visual similarity (e.g., classification), while incorporating task text into retrieval becomes important as question semantics become more diverse (e.g., visual question answering).

\subsection{Retrieval-Augmented In-Context Learning}

The causal and correlational retrieval results are integrated into a unified context for model inference. The final retrieved context is constructed as
\begin{equation}
\mathcal{R} = \mathcal{R}_{\text{causal}} \cup \mathcal{R}_{\text{corr}},
\end{equation}
which is used as context for answer generation:
\begin{equation}
    A_q = \Phi(I_q, Q_q, \mathcal{R}).
\end{equation}
By exposing the model to both correlated and counterfactual instances, CIRCLES encourages reasoning that goes beyond surface-level associations and provides deeper insights into the in-context learning process.
\section{Experiments}

\subsection{Experimental Settings}

\paragraph{Datasets.}
For evaluation, we consider a diverse set of datasets: CUB \cite{wah2011caltech} and Flowers \cite{nilsback2008automated} for fine-grained image classification, OK-VQA \cite{marino2019ok} for open-ended visual question answering, and VizWiz \cite{gurari2018vizwiz} for real-world visual question answering under challenging conditions. For CUB and Flowers, we report classification accuracy (Acc) and weighted F1 scores. For OK-VQA and VizWiz, we use exact match (EM) and word-level F1 metrics. These datasets enable a comprehensive assessment of visual reasoning capabilities across various domains.

\begin{table*}[h!]\small
\centering
\caption{Performance comparison between CIRCLES and baseline methods across different models and datasets. 
Average scores are computed by treating accuracy (Acc) as EM for classification tasks and averaging EM and F1 across all datasets. 
The best results for each model and dataset are highlighted in bold.}
\label{tab:performance_comparison}
\begin{tabular}{ccccccccccccc}
\toprule
\multirow{2.5}{*}{Model} & \multirow{2.5}{*}{Method} 
& \multicolumn{2}{c}{CUB} 
& \multicolumn{2}{c}{Flowers} 
& \multicolumn{2}{c}{OK-VQA} 
& \multicolumn{2}{c}{VizWiz} & \multicolumn{2}{c}{Average}  \\
\cmidrule(lr){3-4} \cmidrule(lr){5-6} \cmidrule(lr){7-8} \cmidrule(lr){9-10} \cmidrule(lr){11-12}
 & & Acc & F1 & Acc & F1 & EM & F1 & EM & F1 & EM & F1 \\
\midrule
\multirow{6}{*}{\makecell{Gemma3\\-4B}}
& None   & 10.56 &  8.19 & 46.76 & 46.50 & 19.12 & 26.70 & 39.85 & 57.26 & 29.07 & 34.66 \\
& Random &  9.98 &  8.10 & 37.01 & 39.51 & 24.81 & 30.74 & 51.84 & 67.08 & 30.91 & 36.36 \\
& RICES  & 65.40 & 67.62 & 86.70 & 87.43 & 26.65 & 32.72 & 56.08 & 70.40 & 58.71 & 64.54 \\
& MUIER  & 65.21 & 67.42 & 86.39 & 87.28 & 26.87 & 32.82 & 56.59 & 70.75 & 58.77 & 64.57 \\
& MMICES & 13.95 & 10.98 & 38.09 & 38.20 & 26.24 & 32.41 & 52.72 & 68.21 & 32.75 & 37.45 \\
& CIRCLES  & \textbf{71.97} & \textbf{72.39} & \textbf{93.32} & \textbf{93.49} & \textbf{31.27} & \textbf{36.89} & \textbf{57.61} & \textbf{71.35} & \textbf{63.54} & \textbf{68.53} \\
\midrule
\multirow{6}{*}{\makecell{Gemma3\\-12B}}
& None   & 30.34 & 25.02 & 67.51 & 64.33 & 25.47 & 33.89 & 56.82 & 71.26 & 45.04 & 48.62 \\
& Random & 29.27 & 25.82 & 71.28 & 71.56 & 33.59 & 39.56 & 70.13 & 80.54 & 51.07 & 54.37 \\
& RICES  & 76.37 & 76.25 & 96.44 & 96.29 & 36.86 & 42.61 & 73.98 & \textbf{83.43} & 70.91 & 74.65 \\
& MUIER  & 76.51 & 76.39 & 96.42 & 96.28 & 36.58 & 42.30 & 73.72 & 83.04 & 70.81 & 74.50 \\
& MMICES & 36.37 & 31.61 & 71.36 & 69.13 & 35.12 & 40.85 & 71.50 & 81.56 & 53.59 & 55.79 \\
& CIRCLES  & \textbf{77.03} & \textbf{76.90} & \textbf{97.77} & \textbf{97.75} & \textbf{37.75} & \textbf{43.23} & \textbf{74.30} & 82.35 & \textbf{71.71} & \textbf{75.06} \\
\midrule
\multirow{6}{*}{\makecell{Qwen2.5\\-VL-3B}} 
& None   &  7.09 &  5.97 & 22.57 & 24.04 & 42.29 & 46.12 & \textbf{75.83} & \textbf{78.58} & 36.95 & 38.68 \\
& Random & 15.81 & 15.63 & 41.76 & 44.57 & 41.54 & 45.30 & 74.48 & 77.40 & 43.40 & 45.73 \\
& RICES  & 72.26 & 73.88 & 93.06 & 92.84 & 42.57 & 46.26 & 70.80 & 73.37 & 69.67 & 71.59 \\
& MUIER  & 72.25 & 73.87 & 93.06 & 92.83 & 41.66 & 45.82 & 71.61 & 74.74 & 69.64 & 71.81 \\
& MMICES & 17.26 & 15.38 & 33.84 & 35.47 & 39.32 & 42.95 & 73.23 & 76.21 & 40.91 & 42.50 \\
& CIRCLES  & \textbf{74.89} & \textbf{76.34} & \textbf{94.70} & \textbf{94.77} & \textbf{43.24} & \textbf{47.27} & 72.93 & 75.33 & \textbf{71.44} & \textbf{73.43} \\
\midrule
\multirow{6}{*}{\makecell{Qwen2.5\\-VL-7B}}
& None   & 14.83 & 11.79 & 43.47 & 42.14 & 33.31 & 38.33 & 68.70 & 73.30 & 40.08 & 41.39 \\
& Random & 26.27 & 24.42 & 47.15 & 49.74 & 41.66 & 46.20 & 76.11 & 78.34 & 47.80 & 49.67 \\
& RICES  & 82.15 & 81.91 & 98.83 & 98.85 & 43.66 & 48.32 & 73.79 & 76.13 & 74.61 & 76.30 \\
& MUIER  & 82.14 & 81.90 & 98.93 & 98.95 & \textbf{44.29} & \textbf{48.76} & 74.67 & 77.08 & 75.01 & 76.67 \\
& MMICES & 34.57 & 29.72 & 49.15 & 47.62 & 43.40 & 47.66 & 70.69 & 72.77 & 49.45 & 49.44 \\
& CIRCLES  & \textbf{82.17} & \textbf{82.13} & \textbf{98.99} & \textbf{99.04} & 43.54 & 48.53 & \textbf{77.63} & \textbf{82.36} & \textbf{75.58} & \textbf{78.02} \\
\bottomrule
\end{tabular}
\end{table*}

\paragraph{Baselines.}
We compare CIRCLES against several in-context learning baselines:
\textbf{None}: zero-shot prompting without in-context examples.
\textbf{Random}: in-context learning with randomly sampled examples.
\textbf{RICES} \cite{yang2022empirical,alayrac2022flamingo}: retrieval based on nearest neighbors using image-image similarity.
\textbf{MUIER} \cite{luo2024does}: multimodal retrieval leveraging both image-image and image-text similarities for example selection.
\textbf{MMICES} \cite{chen2025can}: a two-stage multimodal selector that first retrieves candidates by image-image similarity and then re-ranks them using text-image similarity for improved query alignment.
All methods are implemented with Gemma3 (4B/12B) \cite{team2025gemma} and Qwen2.5-VL (3B/7B) \cite{bai2025qwen2} backbones. CLIP (ViT-g/14) \cite{Radford2021LearningTV} is used as the image/text encoder for retrieval, and the in-context example budget is set to 32 for all methods, where CIRCLES has 16 images in $\mathcal{R}_{\text{causal}}$ and 16 in $\mathcal{R}_{\text{corr}}$. More implementation details are provided in Appendix \ref{app:implementation_details}.

\subsection{Performance Comparison}

Table \ref{tab:performance_comparison} presents a comprehensive evaluation of CIRCLES and baseline methods across multiple datasets and vision-language models (VLMs). The results show that all VLMs benefit from in-context learning, achieving 6.33\% to 118.58\% relative EM improvements over the zero-shot baseline (None). Additionally, retrieval-based methods (RICES, MUIER, MMICES, CIRCLES) consistently outperform random selection, with up to 105.56\% relative improvement, highlighting the importance of selecting relevant examples for effective in-context learning.

Moreover, CIRCLES consistently surpasses other example selection baselines across nearly all datasets and backbone configurations, achieving average relative EM improvements ranging from 0.76\% to 94.02\%. This highlights the effectiveness of composed image retrieval (CIR) in CIRCLES, which complements the similarity-based retrieval with causal understanding. The performance gains are especially pronounced for smaller backbone models such as Gemma3-4B and Qwen2.5-VL-3B, indicating that CIRCLES provides substantial contextual support when the model's internal knowledge is limited.

\begin{figure*}[h!]
    \centering
    \includegraphics[width=1.0\linewidth]{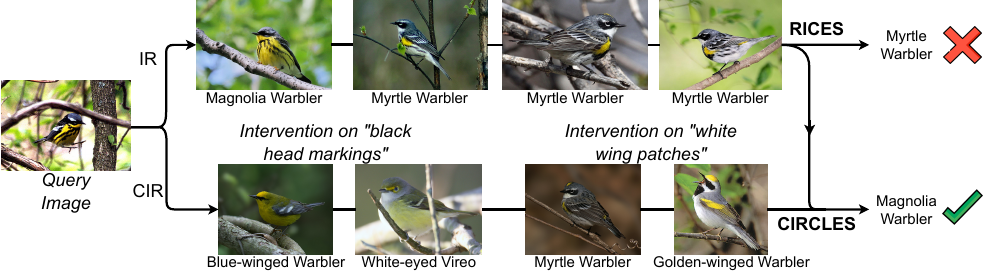}
    \caption{Qualitative comparison of in-context examples retrieved by RICES and CIRCLES for a CUB test image (\textit{Magnolia Warbler}). Top: standard image retrieval (IR) neighbors used by RICES, leading to incorrect predictions. Bottom: counterfactual examples from composed image retrieval (CIR) in CIRCLES, highlighting key attribute changes and guiding the model to the correct label.}
    \label{fig:case_study}
\end{figure*}

Comparing performance improvements across datasets, we find that CIRCLES yields especially pronounced gains on fine-grained classification tasks (CUB, Flowers), where distinguishing subtle attribute differences is critical for accurate predictions. This underscores the value of CIR in scenarios requiring nuanced visual reasoning. While improvements on visual question answering tasks (OK-VQA, VizWiz) are more modest, CIRCLES consistently achieves leading results, demonstrating its effectiveness and versatility across diverse vision-language challenges.

\subsection{Qualitative Analysis of CIRCLES}

Figure \ref{fig:case_study} provides a qualitative comparison of in-context examples retrieved by RICES (images via standard image retrieval) and CIRCLES (images via both standard and composed image retrieval) for a test sample from the CUB dataset. The query image is a \emph{Magnolia Warbler}, whose label is largely determined by key attributes like ``black head markings'' and ``white wing patches''. RICES, which operates purely in the visual similarity space, retrieves a sequence of images that are globally similar to the query (top row). Because these retrieved examples overrepresent Myrtle Warblers, the in-context prompt encourages the model toward the incorrect \emph{Myrtle Warbler} prediction.

In contrast, CIRCLES explicitly constructs counterfactual examples with CIR that intervene on semantically meaningful attributes (bottom row), retrieving images that closely resemble the query but differ in targeted features such as the presence or absence of ``black head markings.'' This approach retrieves counterfactual examples that clearly demonstrate how changes in key attributes result in different bird species labels. By illustrating class transitions across these counterfactuals, CIRCLES provides the VLM with explicit cues about which attributes are decisive, enabling accurate predictions and offering a more interpretable perspective on the in-context learning process.

\subsection{CIRCLES under Information Scarcity}

CIRCLES complements standard image retrieval (IR) with CIR to provide diverse and informative in-context examples, which should be particularly beneficial when the training set contains limited information relevant to the query. To validate this hypothesis, we assess the robustness of CIRCLES and RICES under varying degrees of information scarcity in the training set. Specifically, we simulate different levels of information scarcity by randomly removing a certain percentage of samples from the training set (ranging from 0\% to 75\%) and evaluate the performance of each method on the CUB dataset using different backbone models. The attribute set used in CIRCLES is fixed across all levels of information scarcity to ensure a fair comparison.

\begin{figure}[h!]
    \centering
    \includegraphics[width=1.0\linewidth]{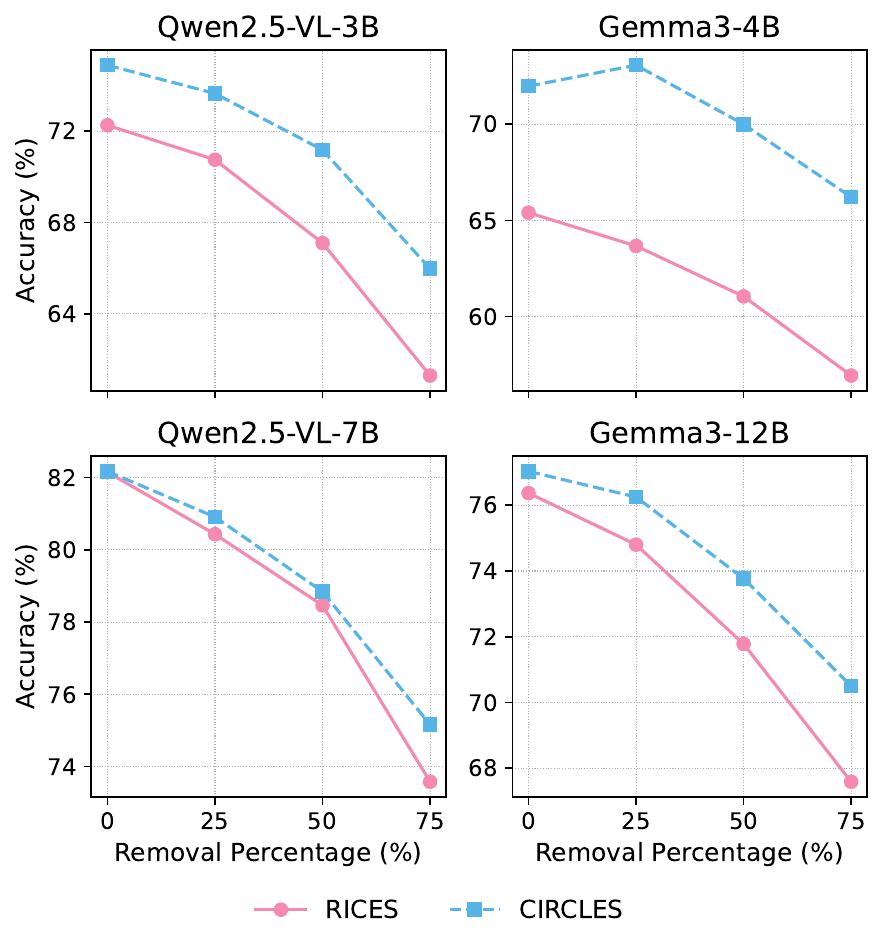}
    \caption{Performance comparison between CIRCLES and RICES on the CUB dataset under varying levels of information scarcity in the training set.}
    \label{fig:robustness}
\end{figure}

As shown in Figure \ref{fig:robustness}, both RICES and CIRCLES experience performance degradation as more training samples are removed. However, CIRCLES consistently outperforms RICES across all levels of information scarcity. Notably, as the percentage of removed training samples increases, CIRCLES's performance advantage over RICES becomes more pronounced. In small-scale VLMs such as Gemma3-4B, the relative improvement of CIRCLES over RICES increases from 10.05\% to 16.28\% when the removal percentage is increased from 0\% to 75\%. For larger models such as Gemma3-12B, the relative improvements are smaller but still show a consistent upward trend, rising from 0.86\% with the full training set to 4.31\% when 75\% of the samples are removed. 

Since CIRCLES shares the same standard image retrieval component as RICES, these results indicate that the CIR component in CIRCLES effectively mitigates the challenges posed by limited relevant information in the training set, enhancing the robustness of in-context learning under information scarcity.

\subsection{Impact of CIR Implementation on CIRCLES} \label{sec:cir_implementation}
To assess the impact of different CIR methods on CIRCLES performance, we compare two representative training-free CIR implementations: CIReVL \cite{karthik2024visionbylanguage} and OSrCIR \cite{tang2025reason}. CIReVL generates counterfactual captions by first producing a caption for the original image and then editing it to modify target attributes, which can sometimes result in generic or less contextually grounded descriptions. In contrast, OSrCIR directly synthesizes captions conditioned on both the query image and manipulation text, enabling more flexible and fine-grained descriptions of attribute changes and their interactions. This approach allows OSrCIR to capture subtle attribute variations and complex dependencies, producing counterfactuals that are more informative and relevant for in-context learning. In our implementation of CIRCLES, we adopt OSrCIR due to its demonstrated superiority in generating high-quality composed image retrievals \cite{tang2025reason}.

\begin{figure}[h!]
    \centering
    \includegraphics[width=1.0\linewidth]{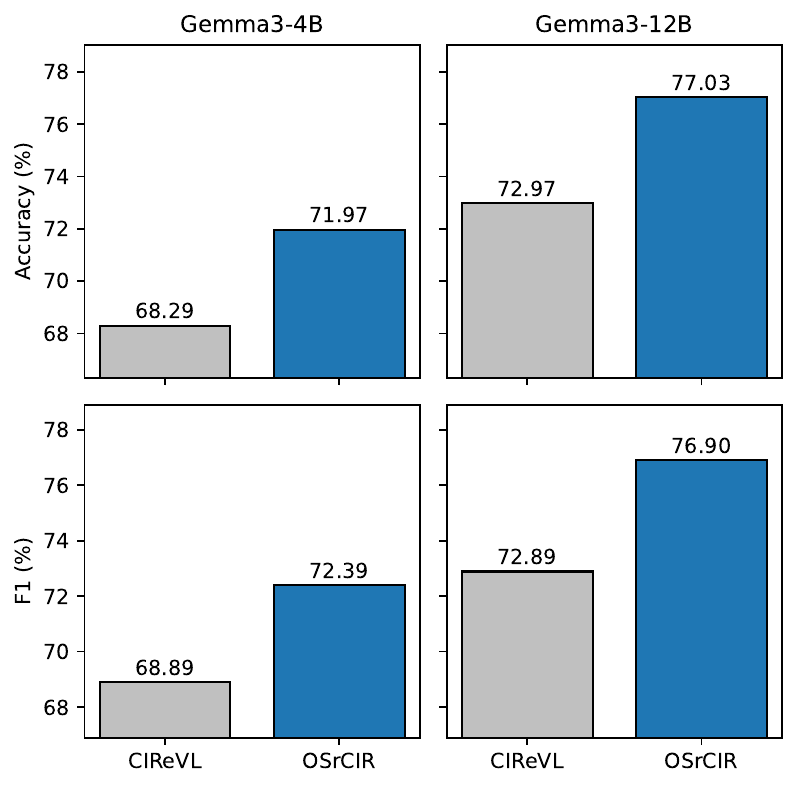}
    \caption{Comparison of CIRCLES performance using CIR implemented by CIReVL and OSrCIR on the CUB dataset with Gemma3 (4B/12B) backbones.}
    \label{fig:cir_method}
\end{figure}

Figure \ref{fig:cir_method} compares CIRCLES instantiated with CIReVL and OSrCIR on the CUB dataset using Gemma3-4B and Gemma3-12B backbones. As shown, CIRCLES with CIR implemented by OSrCIR consistently outperforms CIReVL on both backbone VLMs, achieving relative accuracy improvements ranging from 5.39\% to 5.56\% and relative F1 improvements from 5.08\% to 5.50\%. These results align with the CIR performance comparison reported in \cite{tang2025reason}, where OSrCIR demonstrated superior retrieval accuracy over CIReVL. Overall, the quality of the CIR method directly impacts the effectiveness of CIRCLES, as more accurate and contextually relevant composed image retrievals yield better counterfactual examples for in-context learning.

\begin{figure*}[h!]
    \centering
    \begin{subfigure}[b]{0.255\linewidth}
        \centering
        \includegraphics[width=\linewidth]{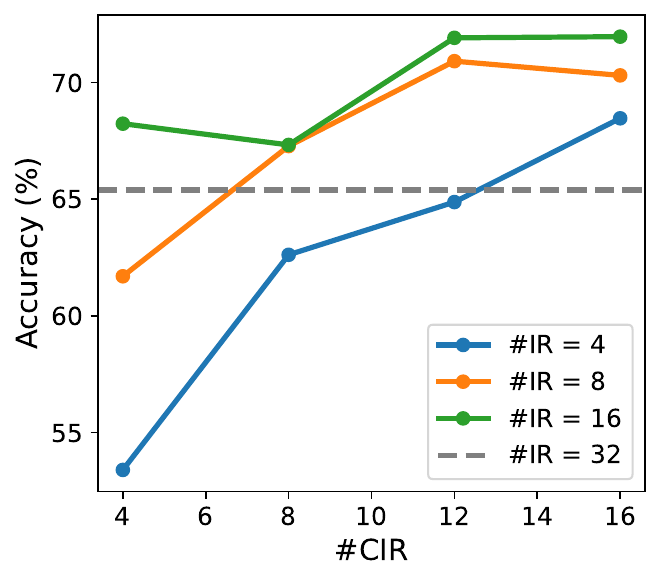}
        \caption{}
        \label{fig:attr_k_scaling}
    \end{subfigure}
    \begin{subfigure}[b]{0.735\linewidth}
        \centering
        \includegraphics[width=\linewidth]{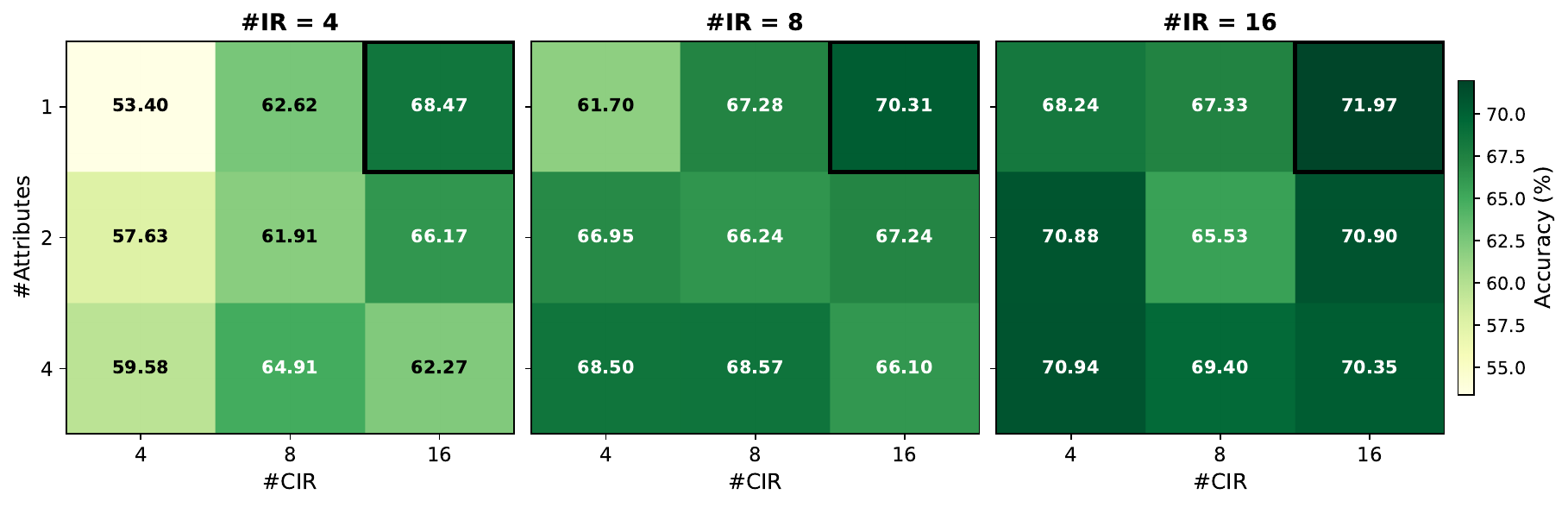}
        \caption{}
        \label{fig:attr_budget}
    \end{subfigure}
    \caption{Retrieval budget analysis on CUB with Gemma3-4B. (a) CIRCLES accuracy vs. number of composed images retrieved (\#CIR) under different standard retrievals (\#IR).
    (b) Accuracy as a function of the number of intervened attributes (\#Attributes) and composed images (\#CIR). 
    }
    \label{fig:combined_attr}
\end{figure*}

As discussed in Section \ref{sec:method}, we further augment CIR with a question-question similarity component to make the retrieval aware of task similarity (Equation \eqref{eq:task_similarity}), which is not present in the original CIR formulation \cite{tang2025reason}. Intuitively, this term biases retrieval toward examples that not only share visual content but also require similar reasoning skills or knowledge types. Because CUB and Flowers have identical questions for all samples, this similarity will not have an effect. Therefore, we evaluate the impact of our proposed enhancement on OK-VQA and VizWiz, where questions are more diverse.

\begin{table}[h!] \small
    \centering
    \caption{Performance comparison of CIRCLES on OK-VQA and VizWiz with (w/) and without (w/o) the question-question similarity component in CIR.}
    \label{tab:cir_task}
    \begin{tabular}{ccccccc}
    \toprule
    \multirow{2.5}{*}{Model} & \multirow{2.5}{*}{Setting} & \multicolumn{2}{c}{OK-VQA} & \multicolumn{2}{c}{VizWiz} \\
    \cmidrule(lr){3-4} \cmidrule(lr){5-6}
     & & EM & F1 & EM & F1 \\
    \midrule
     \multirow{2}{*}{\makecell{Gemma\\-4B}} 
     & w/o & 27.72 & 33.64 & 57.40 & 71.65 \\
     & w/ & {31.27} & {36.89} & 57.61 & 71.35 \\
    \midrule
     \multirow{2}{*}{\makecell{Gemma\\-12B}} 
     & w/o & 33.02 & 39.07 & 74.37 & 83.20 \\
     & w/ & {37.75} & {43.23} & 74.30 & 82.35 \\
    \midrule
     \multirow{2}{*}{\makecell{Qwen\\-3B}} 
     & w/o & 41.12 & 45.33 & 73.28 & 75.94 \\
      & w/ & 43.24 & 47.27 & 72.93 & 75.33 \\
    \midrule
     \multirow{2}{*}{\makecell{Qwen\\-7B}} 
     & w/o & 40.80 & 46.22 & 77.29 & 81.72 \\
     & w/ & 43.54 & 48.53 & 77.63 & 82.36 \\
    \bottomrule
    \end{tabular}
\end{table}

Table \ref{tab:cir_task} compares CIRCLES performance with and without the question-question similarity term. On OK-VQA, we observe consistent improvements across all backbone models, with relative EM gains of +5.16\% to 14.32\% and F1 gains of +4.28\% to 10.65\%. This demonstrates that explicitly matching questions enables retrieval of demonstrations with similar reasoning requirements, which is particularly beneficial for knowledge-intensive, open-ended queries. On VizWiz, the improvements are more modest, likely due to the highly diverse and often noisy user-generated questions and challenging images, where textual similarity is a weaker signal and visual cues are more influential. Nevertheless, the addition of the similarity term does not cause significant performance degradation, suggesting it is a robust default that can yield substantial gains when task structures are more regular, as in OK-VQA.

\subsection{Retrieval Budget Analysis}

We further analyze how to best allocate the in-context example budget between standard image retrieval (IR) and composed image retrieval (CIR) in CIRCLES. Figure \ref{fig:attr_k_scaling} shows classification accuracy as a function of the number of composed images retrieved (\#CIR) for different settings of standard retrievals (\#IR). Increasing \#CIR consistently improves accuracy across all \#IR configurations, highlighting the benefit of CIR for providing targeted and informative context. Importantly, adding composed images yields substantial gains, even when the total number of retrieved examples is lower than using only standard retrieval (grey line: \#CIR=0, \#IR=32). This indicates that CIR enables more efficient use of the retrieval budget, as composed examples focus the model on key attribute-level changes and support causal reasoning. Overall, these results demonstrate that combining CIR and IR leads to more informative in-context examples and improved model performance.

While Figure \ref{fig:attr_k_scaling} examines the effect of varying the number of composed images (\#CIR) when only a single attribute is intervened on (\#Attributes = 1), we further analyze how accuracy changes as both \#Attributes and \#CIR are varied. Figure \ref{fig:attr_budget} presents a heatmap of CIRCLES accuracy on CUB across different hyperparameter settings. When the retrieval budget is limited (e.g., \#CIR = 4), distributing interventions across more attributes per query yields better performance than focusing retrievals on a single attribute. As the budget increases (e.g., \#CIR = 16), allocating more compositions to fewer attributes becomes advantageous. This suggests that the optimal allocation of the in-context example budget shifts from breadth (intervening on more attributes) to depth (focusing on fewer attributes) as \#CIR grows. In practice, these results indicate that under tight budgets, spreading compositions across attributes maximizes coverage, while with larger budgets, concentrating CIR on a smaller subset of attributes provides more precise and informative interventions, because imperfect CIR methods may not always rank the most relevant images at the top.
Further discussions on efficiency and additional design ablations are provided in Appendices \ref{app:efficiency}, \ref{app:attribute_quality}, and \ref{app:ablation_design}.
\section{Conclusion}
We proposed CIRCLES, a retrieval-augmented ICL method that introduces counterfactual demonstrations through attribute-guided composed image retrieval, enabling VLMs to reason beyond surface similarity and better capture the causal structure underlying visual tasks. By integrating compositional interventions with conventional similarity-based retrieval, CIRCLES offers a principled mechanism for enriching demonstration sets with examples that expose disentangled attribute-level variations. Our experiments highlight not only consistent quantitative gains but also qualitative improvements in the diversity and causal informativeness of retrieved examples, particularly under information-scarce regimes where traditional retrieval methods degrade most sharply. These findings suggest that counterfactual retrieval is a practical and effective approach for enhancing visual in-context learning in VLMs.

\section*{Acknowledgments}
This work is supported in part by the US National Science Foundation (NSF) and the National Institute of Health (NIH) under grants IIS-2106913, IIS-2538206, IIS-2529378, CCF-2217071, CNS-2213700, and R01LM014012-01A1. Any recommendations expressed in this material are those of the authors and do not necessarily reflect the views of NIH or NSF.

{
    \small
    \bibliographystyle{ieeenat_fullname}
    \bibliography{main}
}

\appendix
\renewcommand{\thesection}{\Alph{section}}

\clearpage
\setcounter{page}{1}
\maketitlesupplementary

\section{Implementation Details} \label{app:implementation_details}

\subsection{Baseline Implementations}

\paragraph{Backbone vision-language models.}
All baselines and CIRCLES are instantiated with the same backbone vision-language models (VLMs) to ensure a fair comparison. We use Gemma3 \cite{team2025gemma} with 4B and 12B parameters (denoted as Gemma3-4B and Gemma3-12B) and Qwen2.5-VL \cite{bai2025qwen2} with 3B and 7B parameters (denoted as Qwen2.5-VL-3B and Qwen2.5-VL-7B) from the HuggingFace platform. For all models, we run inference with BFloat16 precision using the vLLM library. 

\paragraph{Retrieval backbone.}
For all retrieval-based baselines, including RICES, MUIER, MMICES, and CIRCLES, we use CLIP ViT-g/14 ("laion/CLIP-ViT-g-14-laion2B-s12B-b42K") \cite{Radford2021LearningTV} as the shared image-text encoder. Unless otherwise specified, all visual and textual features are extracted from the CLIP image/text encoder and L2-normalized before similarity computation.

\paragraph{Prompting setup.}
We use a unified prompting template across all methods for each task. For each instance, the prompt includes the query image and question, followed by in-context demonstrations selected by the respective method. The input query is repeated at the end of the prompt for clarity. All methods and VLMs use a decoding temperature of 0.0 for deterministic generation, and the maximum output length is set to 512 tokens for all experiments. For VQA tasks (OK-VQA and VizWiz), in-context demonstrations consist of the original question-answer pairs. For classification tasks (CUB and Flowers), we use fixed question templates and the corresponding class labels as in-context examples. Specifically, the question template for CUB is ``What is the category of the bird in this image?'', and for Flowers, ``What is the category of the flower in this image?''.

\paragraph{Baselines.}

We implement the following baselines:

\begin{itemize}
\item \textbf{None} (zero-shot): The backbone VLM receives only the task description and query example, without any in-context demonstrations.
\item \textbf{Random}: In-context demonstrations are uniformly sampled from the training set for each query, without retrieval.
\item \textbf{RICES} \cite{yang2022empirical,alayrac2022flamingo}: For each query, CLIP image embeddings are computed and the top-$K$ nearest neighbors in the visual feature space are retrieved by cosine similarity. Retrieved examples are sorted by similarity and added to the prompt.
\item \textbf{MUIER} \cite{luo2024does}: Following the original paper, we use a multimodal similarity score combining image-image and image-text similarities, with the same CLIP backbone as RICES. The image-text similarity is computed between the query image and each candidate example's question text. Candidates are ranked by the multimodal score, and the top-$K$ are selected as demonstrations.
\item \textbf{MMICES} \cite{chen2025can}: MMICES is implemented as a two-stage selector. First, 1024 candidate demonstrations are retrieved using image-image similarity. Second, these candidates are re-ranked by a text-image similarity score measuring how well each candidate's image matches the query's question. The top-$K$ re-ranked examples are used as in-context demonstrations.
\end{itemize}

All baselines are evaluated with a fixed in-context budget of $K=32$ demonstrations per query, unless otherwise specified in ablations. Demonstrations are selected from the training set of each dataset.

\subsection{Implementation of CIRCLES}

CIRCLES augments standard image retrieval (IR) with composed image retrieval (CIR) to construct richer and more causally informative in-context examples. For each query, CIRCLES first retrieves visually similar neighbors (IR) using CLIP, and then invokes a training-free CIR module to generate counterfactual examples via language-guided attribute interventions. The final in-context set is composed of both IR and CIR examples, constrained by a fixed total budget of 32 examples unless otherwise stated.

\paragraph{Attribute Extraction.}
For each query, we prompt the backbone VLM to identify the most prominent attributes visible in the image that are relevant to the given question. The VLM outputs a ranked list of attributes, ordered from most to least important for answering the query. This approach does not rely on explicit attribute labels or fixed vocabularies; instead, attribute phrases are generated dynamically by the VLM based on the image and question context.

\paragraph{Standard image retrieval (IR).}
The IR component in CIRCLES is identical to RICES: CLIP image embeddings are computed for all training images and the query image, and we retrieve the top-$K_{IR}$ neighbors by cosine similarity. For a full in-context budget of 32 examples, we set $K_{IR}$=16 and allocate the remaining budget to CIR (e.g., $K_{CIR}$=16).

\paragraph{Composed image retrieval (CIR) with OSrCIR.}
We implement CIR using OSrCIR \cite{tang2025reason}, a training-free framework that synthesizes captions conditioned on the query image and attribute-manipulation text. For each selected attribute of a query image, we prompt the VLM to generate a counterfactual caption describing the image with the desired attribute change. This composed caption is encoded with CLIP and used to retrieve images from the training set whose visual embeddings best match the description. As detailed in Section \ref{sec:method}, candidates are ranked by the sum of image-image and text-text similarity scores, both normalized to the range $[0, 1]$ via L2-normalization. The top $K_{CIR}$ candidates are selected as CIR examples and combined with IR examples to form the final in-context set.

The detailed prompts used in our implementation are provided in Appendix \ref{app:template}.

\section{Discussions on Efficiency} \label{app:efficiency}

Compared to retrieval-only in-context learning baselines such as RICES, MUIER, and MMICES, CIRCLES introduces additional computation at inference time due to attribute extraction and composed image retrieval. For each query, we first invoke the VLM once to identify salient attributes, and then issue one VLM call per selected attribute to generate a counterfactual caption describing the desired manipulation. These calls are in addition to the final answer-generation call, which is shared by all methods. In contrast, standard retrieval baselines perform only CLIP-based retrieval followed by a single VLM call for answer generation.

Although CIRCLES introduces additional VLM calls for attribute extraction and composed retrieval, these calls use short prompts, and the dominant cost still comes from the final in-context inference step with 32 demonstrations. To quantify this overhead, Table \ref{tab:token_usage} reports the average number of tokens processed per question for RICES and CIRCLES across datasets. Overall, CIRCLES adds only about 10\% token overhead relative to RICES, while providing consistently better performance in the main experiments.

\begin{table}[h!]
\small
\centering
\caption{Average token usage per question for RICES and CIRCLES. The additional VLM calls in CIRCLES introduce only a modest overhead (around 10\%).}
\label{tab:token_usage}
\resizebox{\linewidth}{!}{%
\begin{tabular}{cccccc}
\toprule
Model & Method & CUB & Flowers & OK-VQA & VizWiz \\
\midrule
\multirow{2}{*}{\makecell{Gemma3\\-4B}}
& RICES   & 11.1k & 10.0k & 9.4k  & 10.2k \\
& CIRCLES & 12.2k & 11.2k & 10.6k & 11.4k \\
\midrule
\multirow{2}{*}{\makecell{Qwen2.5\\-VL-3B}}
& RICES   & 9.9k  & 15.8k & 12.8k & 33.3k \\
& CIRCLES & 10.9k & 17.2k & 14.1k & 35.9k \\
\bottomrule
\end{tabular}
}
\end{table}

To further compare CIRCLES with baseline methods under the same VLM call budget, we test a compute-matched variant of RICES, denoted RICES*, which performs multiple answer generations ($K=3$) followed by self-consistency. As shown in Table \ref{tab:compute_matched}, CIRCLES consistently outperforms RICES* across datasets and model scales. This suggests that the improvements of CIRCLES are not explained solely by additional generations, but by the quality of the attribute-guided retrieved demonstrations.

\begin{table}[h!]
\small
\centering
\caption{Compute-matched comparison between CIRCLES and RICES*. RICES* uses multiple generations ($K=3$) with self-consistency under the same inference-time call budget.}
\label{tab:compute_matched}
\resizebox{\linewidth}{!}{%
\begin{tabular}{cccccc}
\toprule
Model & Method & CUB & Flowers & OK-VQA & VizWiz \\
\midrule
\multirow{2}{*}{\makecell{Gemma3\\-4B}}
& RICES*  & 64.73 & 86.60 & 26.34 & 56.17 \\
& CIRCLES & 71.97 & 93.32 & 31.27 & 57.61 \\
\midrule
\multirow{2}{*}{\makecell{Gemma3\\-12B}}
& RICES*  & 75.70 & 96.02 & 36.64 & 74.09 \\
& CIRCLES & 77.03 & 97.77 & 37.75 & 74.30 \\
\bottomrule
\end{tabular}
}
\end{table}

Finally, we view CIRCLES as a first, training-free instantiation of counterfactual retrieval rather than an efficiency-optimized endpoint. Our current implementation relies on an LLM to generate counterfactual captions, which are then encoded with CLIP and used for composed retrieval. As composed image retrieval models and multimodal embedding architectures mature, one could instead directly embed joint (image, manipulation text) queries, or precompute attribute-aware embeddings, thereby reducing or even eliminating the need for multiple online LLM calls. Such designs would bring the computational profile of counterfactual retrieval closer to that of standard similarity-based retrieval, while retaining the robustness and causal benefits demonstrated by CIRCLES.

\section{Robustness and Sensitivity}

Our main experiments use deterministic decoding with temperature 0.0. To assess robustness under stochastic generation, we additionally evaluate RICES and CIRCLES with Gemma3-4B using temperature 1.0 and repeat decoding five times. \Cref{tab:variance} reports the mean and standard deviation across the benchmark datasets.

\begin{table}[h!]\small
\centering
\caption{Robustness under stochastic decoding for Gemma3-4B. We set temperature = 1.0, repeat decoding five times, and report mean $\pm$ standard deviation.}
\label{tab:variance}
\begin{tabular}{lcccc}
\toprule
 Method & CUB & Flowers & OK-VQA & VizWiz \\
\midrule
 RICES   & \makecell{64.42\\\scriptsize$\pm$0.32} & \makecell{86.42\\\scriptsize$\pm$0.24} & \makecell{26.08\\\scriptsize$\pm$0.23} & \makecell{55.69\\\scriptsize$\pm$0.07} \\
 \midrule
 CIRCLES & \makecell{70.30\\\scriptsize$\pm$0.28} & \makecell{92.87\\\scriptsize$\pm$0.13} & \makecell{30.54\\\scriptsize$\pm$0.21} & \makecell{57.04\\\scriptsize$\pm$0.23} \\
\bottomrule
\end{tabular}
\end{table}

\begin{figure*}[h!]
    \centering
    \includegraphics[width=1\linewidth]{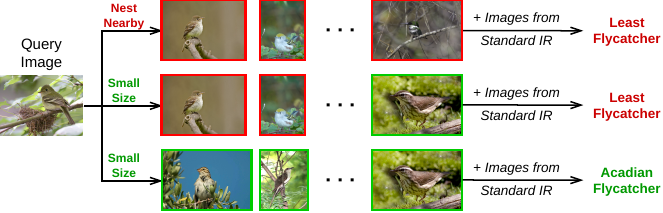}
    \caption{Qualitative examples of CIRCLES. Top: failure due to a non-discriminative extracted attribute. Middle: failure due to weak composed retrieval. Bottom: success case with better matched composed examples. Red denotes unhelpful retrieved examples or incorrect predictions, and green denotes helpful retrieved examples or the correct prediction.}
    \label{fig:failure_cases}
\end{figure*}

CIRCLES consistently outperforms RICES under stochastic decoding on all four datasets. Moreover, the standard deviations are small relative to the performance gaps between the two methods, suggesting that the gains from CIRCLES are stable across repeated stochastic runs rather than driven by favorable sampling variation.

\Cref{fig:failure_cases} shows two representative failure modes of CIRCLES and one successful case. In the top row, the extracted attribute (\emph{nest nearby}) is visually plausible but not discriminative, so the composed examples provide little useful signal beyond standard IR. In the middle row, the extracted attribute (\emph{small size}) is relevant, but the composed retrieval is only weakly aligned with the target, limiting its benefit for prediction. In the bottom row, the same attribute yields better matched composed examples and leads to the correct answer. These cases suggest that the effectiveness of CIRCLES depends on both reliable attribute extraction and high-quality attribute-conditioned retrieval, which we view as promising directions for future improvement.

\section{Effect of Attribute Quality and Counterfactual Retrieval} \label{app:attribute_quality}

We first investigate how the quality of attribute information affects the performance of CIRCLES. On CUB, in addition to the attributes extracted by the backbone VLM, we have access to class-specific attribute frequency tables provided by the dataset. For each attribute, we compute a discriminativeness score as the gap between its frequency within a given class and its highest frequency across all other classes, and rank attributes for each class by this score. Then, for a query image, we form a sample-specific attribute list by taking the top-ranked attributes for its class and pruning those that are not annotated as present in the image. This provides a set of clean, image-level attribute descriptions that we feed to CIRCLES in place of VLM-extracted attributes. As shown in Table \ref{tab:cub_gt}, using these ground-truth attributes yields consistent improvements over VLM-extracted attributes for both Gemma3-4B and Gemma3-12B, confirming that accurate attribute identification is beneficial for composed image retrieval and, consequently, for in-context learning. At the same time, the gains are relatively modest, suggesting that the backbone VLM is already able to recover attributes that are close to the dataset oracle.

\begin{table}[h!]\small
\centering
\caption{Effect of replacing VLM-extracted attributes with ground-truth attribute annotations on CUB.}
\label{tab:cub_gt}
\begin{tabular}{ccccccccccccc}
\toprule
\multirow{2.5}{*}{\makecell{Attribute\\Source}}
& \multicolumn{2}{c}{Gemma3-4B} & \multicolumn{2}{c}{Gemma3-12B} \\
\cmidrule(lr){2-3} \cmidrule(lr){4-5}
 & Acc & F1 & Acc & F1  \\
\midrule
LLM & 71.97 & 72.39 & 77.03 & 76.90 \\
Dataset & 72.44 & 72.82 & 77.65 & 77.46 \\
\bottomrule
\end{tabular}
\end{table}

Next, we explore whether it is the explicit use of attributes or the counterfactual retrieval itself that drives the gains of CIRCLES. To disentangle these factors, we compare three settings on all four benchmarks: (1) standard image retrieval only (IR), which corresponds to the RICES-style baseline; (2) IR with attribute information but without CIR (IR+Attr), where we retrieve images using IR and append the extracted attributes as additional textual context without composing counterfactual queries; and (3) IR with full CIR (IR+CIR), which is our proposed CIRCLES setting where attributes are used to generate counterfactual captions and retrieve manipulated visual examples.

\begin{table*}[h!] \small
\centering
\caption{Ablation of CIR versus attribute-only prompting on Gemma3-4B. IR denotes standard image retrieval; Attr denotes adding textual attributes without CIR.}
\label{tab:attr_only}
\begin{tabular}{ccccccccccccc}
\toprule
\multirow{2.5}{*}{Setting} 
& \multicolumn{2}{c}{CUB} 
& \multicolumn{2}{c}{Flowers} 
& \multicolumn{2}{c}{OK-VQA} 
& \multicolumn{2}{c}{VizWiz}  \\
\cmidrule(lr){2-3} \cmidrule(lr){4-5} \cmidrule(lr){6-7} \cmidrule(lr){8-9} 
& Acc & F1 & Acc & F1 & EM & F1 & EM & F1 \\
\midrule
IR &  65.40 & 67.62 & 86.70 & 87.43 & 26.65 & 32.72 & 56.08 & 70.40 \\
IR + Attr & 71.99 & 71.91 & 92.91 & 93.53 & 26.16 & 32.63 & 53.83 & 68.82 \\
IR + CIR  & \textbf{71.97} & \textbf{72.39} & \textbf{93.32} & \textbf{93.49} & \textbf{31.27} & \textbf{36.89} & \textbf{57.61} & \textbf{71.35} \\
\bottomrule
\end{tabular}
\end{table*}

The results for Gemma3-4B are summarized in Table \ref{tab:attr_only}. We observe that providing attribute information alone (IR+Attr) substantially improves performance over IR on fine-grained classification tasks such as CUB and Flowers, where attribute recognition is central to the task. In contrast, on open-ended VQA benchmarks like OK-VQA and VizWiz, simply adding attributes without counterfactual retrieval slightly degrades performance, likely because the added textual descriptions can bias the model toward spurious cues without offering new visual evidence. In all cases, enabling CIR on top of IR and attributes (IR+CIR) yields the best performance, with clear gains over both IR and IR+Attr across all datasets. This suggests that attributes are most effective when they are grounded through counterfactual visual examples, which help the model better interpret and utilize fine-grained attribute cues in diverse downstream tasks.

\section{More Ablations on CIRCLES Designs} \label{app:ablation_design}

While Section \ref{sec:cir_implementation} ablates several design choices within the CIR module itself, here we further probe the overall CIRCLES framework by examining (i) different implementations of the underlying image retrieval (IR) component and (ii) the relative contributions of IR and CIR.

\paragraph{Alternative IR similarity functions.}
In the main experiments, we implement IR using image-image similarity in CLIP space, following the RICES design. To understand whether richer similarity measures can further help CIRCLES, we consider two additional variants inspired by prior work. First, we augment image-image similarity with image-text similarity as in MUIER, where we also match the query image against the question text associated with each candidate example. Second, we combine image-image similarity with the text-text similarity used in our CIR implementation to capture task similarity between questions. Since CUB and Flowers are classification tasks with an identical question template for all test images, text-based similarity is uninformative there. We therefore restrict these IR ablations to OK-VQA and VizWiz.

\begin{table}[h!]\small
\centering
\caption{CIRCLES with different implementations of the image retrieval component (Gemma3-4B).}
\label{tab:ir_ablation}
\begin{tabular}{ccccccccccccc}
\toprule
\multirow{2.5}{*}{Similarity} 
& \multicolumn{2}{c}{OK-VQA} 
& \multicolumn{2}{c}{VizWiz} \\
\cmidrule(lr){2-3} \cmidrule(lr){4-5}
& EM & F1 & EM & F1 \\
\midrule
image-image & 31.27 & 36.89 & 57.61 & 71.35 \\
image-image + image-text  & 30.14 & 35.67 & 57.81 & 71.63  \\
image-image + text-text  & 31.37 & 36.74 & 59.09 & 72.64 \\
\bottomrule
\end{tabular}
\end{table}

Results in Table \ref{tab:ir_ablation} show that simply adding image-text similarity does not improve performance, and in fact slightly degrades results on OK-VQA, mirroring the small gap between RICES and MUIER observed in Table \ref{tab:performance_comparison}. In contrast, incorporating text-text similarity on top of image-image similarity yields small but consistent gains on both datasets, especially on VizWiz. This suggests that, for open-ended VQA, capturing task similarity at the text level is beneficial, while the simple image-image retrieval used in our main CIRCLES configuration is already a strong baseline.

\paragraph{IR versus CIR versus IR+CIR.}
We also disentangle the effects of IR and CIR by comparing three variants: (1) \emph{IR only}, which corresponds to a standard retrieval-based ICL setup using only retrieved images and their labels; (2) \emph{CIR only}, where we discard the original retrieved examples and retain only the counterfactual examples produced by CIR; and (3) \emph{IR + CIR} (full CIRCLES), which includes both the original retrieved examples and their counterfactual counterparts in the in-context prompt. The results on Gemma3-4B are summarized in Table \ref{tab:ir_cir_ablation}.

\begin{table*}[h!] \small
\centering
\caption{Ablation of IR and CIR in CIRCLES, tested on Gemma3-4B.}
\label{tab:ir_cir_ablation}
\begin{tabular}{ccccccccccccc}
\toprule
\multirow{2.5}{*}{Setting} 
& \multicolumn{2}{c}{CUB} 
& \multicolumn{2}{c}{Flowers} 
& \multicolumn{2}{c}{OK-VQA} 
& \multicolumn{2}{c}{VizWiz}  \\
\cmidrule(lr){2-3} \cmidrule(lr){4-5} \cmidrule(lr){6-7} \cmidrule(lr){8-9} 
& Acc & F1 & Acc & F1 & EM & F1 & EM & F1 \\
\midrule
IR only & 65.40 & 67.62 & 86.70 & 87.43 & 26.65 & 32.72 & 56.08 & 70.40 \\
CIR only  & 25.16 & 26.14 & 60.24 & 65.85 & 29.81 & 35.52 & 54.06 & 68.29 \\
IR + CIR  & \textbf{71.97} & \textbf{72.39} & \textbf{93.32} & \textbf{93.49} & \textbf{31.27} & \textbf{36.89} & \textbf{57.61} & \textbf{71.35} \\
\bottomrule
\end{tabular}
\end{table*}

On fine-grained classification benchmarks (CUB and Flowers), CIR only lags far behind IR only, indicating that counterfactual examples alone are not sufficient to capture the subtle visual prototypes needed for class recognition. In contrast, on OK-VQA and VizWiz, CIR only achieves performance comparable to IR only, slightly improving EM and F1 on OK-VQA while being close on VizWiz. Across all four datasets, however, the combined IR + CIR variant consistently yields the best performance, sometimes by a large margin (e.g., +6-7 accuracy points over IR only on CUB and Flowers). These trends highlight that IR and CIR play complementary roles: IR provides realistic, prototypical examples that anchor the model’s understanding of the task, while CIR introduces targeted counterfactual variations that clarify the role of key attributes and reduce spurious correlations. Together, they enable CIRCLES to leverage both factual and counterfactual experience for more robust in-context learning.

\section{Additional Results on Benchmarks and Model Scaling}

\paragraph{Results on ScienceQA.}
To further examine the generality of CIRCLES beyond image classification and visual question answering tasks, we additionally evaluate CIRCLES on ScienceQA \cite{lu2022learn}, a more challenging multimodal reasoning benchmark. Following the same in-context learning protocol used in the main paper, we construct the demonstration pool from the ScienceQA validation split and report performance on the test split.
Table \ref{tab:scienceqa} shows that CIRCLES consistently outperforms all compared example selection baselines on ScienceQA. The gains are modest but consistent across both model sizes. 

\begin{table}[h!]
\centering
\small
\caption{Test accuracy (\%) on ScienceQA. In-context examples are retrieved from the validation split. The experiments are run on the Qwen2.5-VL model family.}
\label{tab:scienceqa}
\begin{tabular}{lcccc}
\toprule
Size & Random & RICES & MUIER & CIRCLES \\
\midrule
3B & 78.33 & 80.12 & 80.66 & \textbf{80.71} \\
7B & 85.18 & 88.05 & 87.90 & \textbf{88.35} \\
\bottomrule
\end{tabular}
\end{table}

\paragraph{Scaling to Larger Models: Gemma3-27B.}
To examine whether the gains of CIRCLES persist at a larger model scale, we additionally evaluate Gemma3-27B, a larger model in the Gemma3 family, on the same four benchmarks used in the main paper. Table \ref{tab:gemma27b} shows that CIRCLES continues to outperform RICES across all datasets, indicating that the benefit of attribute-conditioned, contrastive retrieval is not limited to smaller or mid-sized backbones.

\begin{table}[h!]
\centering
\small
\caption{Performance comparison on Gemma3-27B.}
\label{tab:gemma27b}
\begin{tabular}{lcccc}
\toprule
Method & CUB & Flowers & OKVQA & VizWiz \\
\midrule
RICES   & 69.76 & 93.98 & 38.66 & 67.24 \\
CIRCLES & \textbf{72.21} & \textbf{97.51} & \textbf{39.14} & \textbf{68.72} \\
\bottomrule
\end{tabular}
\end{table}

\section{Prompt Templates} \label{app:template}

We provide the detailed prompt templates used for each dataset in this section. Figure \ref{fig:prompt_attr} shows the template used to extract key visual attributes that are relevant for answering the question. Given an input image and question, the VLM is instructed to list a small set of concise attributes, which are then used as the basis for constructing counterfactual manipulations. Figure \ref{fig:prompt_caption} contains the template used to generate counterfactual captions with targeted attribute changes. The system prompt in Figure \ref{fig:prompt_caption} follows the one used in OSrCIR~\cite{tang2025reason}.

Figures \ref{fig:prompt_none}--\ref{fig:prompt_circles} show the templates used for VLM inference under different in-context learning settings: Figure \ref{fig:prompt_none} corresponds to the setting without in-context examples (``None'' in Table \ref{tab:performance_comparison}); Figure \ref{fig:prompt_rices} illustrates the baseline prompt used for in-context learning methods such as RICES; and Figure \ref{fig:prompt_circles} presents the full CIRCLES prompt, which augments the standard retrieved demonstrations with the counterfactual examples produced by CIR. Across all of these templates, the value of \verb|{{Task Type}}| is set to ``Image Classification'' for CUB and Flowers, and to ``Visual Question Answering'' for OK-VQA and VizWiz, so that the VLM is explicitly informed of the task format.

For the classification datasets (CUB and Flowers), we additionally enforce a closed set of answer options to align with the standard evaluation protocol. Concretely, for these datasets, we append the sentence ``You need to choose one of the following options: \verb|{{Options}}|'' immediately after the first sentence describing the task type, where \verb|{{Options}}| is replaced by the list of candidate class names for the given example. For OK-VQA and VizWiz, which are evaluated in an open-ended manner, we do not provide such options and instead allow the model to freely generate answers conditioned on the image, question, and in-context demonstrations.

\begin{figure*}[h!]
\begin{center}
\begin{AIbox}{Prompt template for attribute extraction}
Identify the key attributes of the following image that are most relevant to answering the question.\\

\verb|{{Image}}|\\
Question: \verb|{{Question}}|\\

Please list the top \verb|{{num_attributes}}| key attributes as short phrases in a section named `\#\#\# Attributes', one per line, ordered from most to least important.

\end{AIbox}
\end{center}
\caption{Prompt template for attribute extraction.}
\label{fig:prompt_attr}
\end{figure*}
\begin{figure*}[h!]
\begin{center}
\begin{AIbox}{Prompt template for generating counterfactual caption}

\verb|{{System Prompt by OSrCIR}}|\\

\verb|{{Image}}|\\

Manipulation Text: Change the attribute \verb|{{Attribute}}| to a different plausible value. Ensure the modified caption is concise and contains no more than 77 tokens.
\end{AIbox}
\end{center}
\caption{Prompt template for generating counterfactual caption based on the query image and identified key attribute.}
\label{fig:prompt_caption}
\end{figure*}
\begin{figure*}[h!]
\begin{center}
\begin{AIbox}{Prompt template for VLM inference used by None}
Your task is to perform \verb|{{Task Type}}|.\\

\verb|{{Image}}|\\
Question: \verb|{{Question}}|\\

Please provide your response by directly outputting the answer.
\end{AIbox}
\end{center}
\caption{Prompt template for VLM inference used by None (zero-shot learning).}
\label{fig:prompt_none}
\end{figure*}
\begin{figure*}[h!]
\begin{center}
\begin{AIbox}{Prompt template for VLM inference used by Random/RICES/MUIER/MMICES}
Your task is to perform \verb|{{Task Type}}|.\\

\verb|{{Image}}|\\
Question: \verb|{{Question}}|\\

Here are \verb|{{K_IR}}| in-context examples to help you answer the question:\\

\verb|{{Retrieved Image}}|\\
Question: \verb|{{Retrieved Question}}|\\
Answer: \verb|{{Retrieved Answer}}|\\

\verb|{{Retrieved Image}}|\\
......\\

Here is the original question again.\\
\verb|{{Image}}|\\
Question: \verb|{{Question}}|\\

Please provide your response by directly outputting the answer.
\end{AIbox}
\end{center}
\caption{Prompt template for VLM inference used by Random, RICES, MUIER, and MMICES.}
\label{fig:prompt_rices}
\end{figure*}
\begin{figure*}[h!]
\begin{center}
\begin{AIbox}{Prompt template for VLM inference used by CIRCLES}
Your task is to perform \verb|{{Task Type}}|.\\

\verb|{{Image}}|\\
Question: \verb|{{Question}}|\\

Here are \verb|{{K_IR}}| in-context examples to help you answer the question:\\

\verb|{{Retrieved Image}}|\\
Question: \verb|{{Retrieved Question}}|\\
Answer: \verb|{{Retrieved Answer}}|\\

\verb|{{Retrieved Image}}|\\
......\\

Examples retrieved based on the target image description after changing \verb|{{Attribute}}| (caption: \verb|{{Caption}}|):\\

\verb|{{Retrieved Image}}|\\
Question: \verb|{{Retrieved Question}}|\\
Answer: \verb|{{Retrieved Answer}}|\\

\verb|{{Retrieved Image}}|\\
......\\

Here is the original question again.\\
\verb|{{Image}}|\\
Question: \verb|{{Question}}|\\

Please provide your response by directly outputting the answer.
\end{AIbox}
\end{center}
\caption{Prompt template for VLM inference used by CIRCLES.}
\label{fig:prompt_circles}
\end{figure*}

\end{document}